RESEARCH ARTICLE

# Reinforcement Learning Aided Sequential Optimization for Unsignalized Intersection Management of Robot Traffic

NISHCHAL HOYSAL G[1], (Student Member, IEEE),
AND PAVANKUMAR TALLAPRAGADA[1,2], (Member, IEEE)
[1]Robert Bosch Centre for Cyber Physical Systems, Indian Institute of Science at Bengaluru, Bengaluru 560012, India
[2]Department of Electrical Engineering, Indian Institute of Science at Bengaluru, Bengaluru 560012, India

Corresponding author: Nishchal Hoysal G (nishchalg@iisc.ac.in)

This work was supported in part by Nokia Corporate Social Responsibility (CSR) grant through the Nokia Centre of Excellence in Networked Robotics at Indian Institute of Science (IISc).

**ABSTRACT** We consider the problem of optimal unsignalized intersection management, wherein we seek to obtain safe and optimal trajectories, for a set of robots that arrive randomly and continually. This problem involves repeatedly solving a mixed integer program (with robot acceleration trajectories as decision variables) with different parameters, for which the computation time using a naive optimization algorithm scales exponentially with the number of robots and lanes. Hence, such an approach is not suitable for real-time implementation. In this paper, we propose a solution framework that combines learning and sequential optimization. In particular, we propose an algorithm for learning a shared policy that given the traffic state information, determines the crossing order of the robots. Then, we optimize the trajectories of the robots sequentially according to that crossing order. This approach inherently guarantees safety at all times. We validate the performance of this approach using extensive simulations and compare our approach against 5 different heuristics from the literature in 9 different simulation settings. Our approach, on average, significantly outperforms the heuristics from the literature in various metrics like objective function, weighted average of crossing times and computation time. For example, in some scenarios, we have observed that our approach offers up to 150% improvement in objective value over the first come first serve heuristic. Even on untrained scenarios, our approach shows a consistent improvement (in objective value) of more than 30% over all heuristics under consideration. We also show through simulations that the computation time for our approach scales linearly with the number of robots (assuming all other factors are constant). We further implement the learnt policies on physical robots with a few modifications to the solution framework to address real-world challenges and establish its real-time implementability.

## I. INTRODUCTION
Unsignalized intersection management [1] requires that a number of robots coordinate their trajectories for ensuring safe and efficient use of the intersection. This problem and its parts have been studied under various names like Cooperative intersection management [1], Intersection management of CAVs [2], Coordination of CAVs at intersection [3], [4], Cooperative intersection control/crossing [5], [6], [7], Coordination at unsignalized intersections [8], Autonomous intersection management [9] etc. Its application can be found in contexts like automated warehouses with hundreds or thousands of mobile robots. The problem of optimal unsignalized intersection management involves getting optimal and safe

The associate editor coordinating the review of this manuscript and approving it for publication was Zhiwu Li.

trajectories for the considered robots to cross the intersection. Note that optimal trajectories inherently determine an optimal crossing order. Solution methods to such a problem inherently involve repeatedly solving mixed integer programs, for which the computational complexity of naive optimization methods scales very badly with the number of robots and lanes. As a result, they are not suitable for real-time implementation. In this work, we propose a learning based solution framework to get safe, near-optimal solutions in real-time, a combination that is not addressed in any single framework in the literature.

### A. RELATED WORK

Unsignalized intersection management for robots or connected and automated vehicles has been studied using different methods and tools over the years. Surveys in [1], [2], [10], [11], and [12] give a detailed description of the recent literature. Here we focus on the optimal unsignalized intersection management problem, which is a very common formulation in the area. While some formulate this problem as a mixed integer linear program [3], [13], [14], [15], others consider a model predictive control approach [4] and some formulate a non-linear problem and use genetic algorithms to solve it [5], [6]. While typically the optimization goal is to minimize the cumulative travel times or maximize the cumulative distance covered by all the robots, recent works like [7], [16], [17], and [18] also take energy consumption into account while computing the optimal trajectories. Works like [9] and [19] discretize the space and use $A^*$ like algorithms to get collision free optimal trajectories for the participating agents.

As naive or generic optimization methods for these problems scale very badly with the number of robots and lanes, there is also interest in developing computationally simpler solution methods. With this motivation, [8], [20], [21], [22] propose solutions that cluster the vehicles/robots in order to reduce the computational or communication effort. To address computational complexity, works like [23] describe general ideas of prioritized motion planning for coordination among multiple agents, but, assuming the knowledge of some pre-assigned priorities to each agent. Similarly, in line with prioritized planning idea, some other works, such as [14] and [24] use model based heuristics to decide the crossing order and solve centralized/decentralized optimization problems preserving that order to get the trajectories for involved robots in an intersection. Work [25] uses a similar approach for the ramp merging problem. While some works like [26], [27] consider finding optimal and safe trajectories to robots assuming a priority (e.g. first-in first-out) rule, some like [28], [29] propose an optimization problem to get such priority rule and solve another optimization problem (in receding horizon control framework) to get trajectories. Work [30] proposes a two-stage optimization method combining the discrete tile and conflict point methods. In [31], optimization based methods are proposed for intersection management with bounded location uncertainties of agents. Other approaches to the problem include auction based methods [32], [33] and first come first serve based reservation of regions against time [19]. Work in [33] proposes a game-theoretic social-welfare optimal auction strategy to decide the crossing order. Others [4], [34] construct/use a priority graph to get feasible solutions and conflict resolution. Work [35] proposes an algorithm which improves upon the reservation based strategy. Work [18] uses similar approach to the intersection management problem and computes the priority/crossing order using a set of features associated with the robots. Then the robots solve an optimal control problem sequentially to obtain their trajectories.

Some works like [36], [37], and [38] take a new approach and use gaussian process to get multi-agent trajectories and motion plans. Specifically, [36] and [37] use factor graph methods to improve computation time.

The current literature on the use of learning methods for intersection management includes [39], [40], [41], [42], [43], [44], [45]. These papers consider only a first order kinematic model for the robots and the learnt policy gives the position trajectories of the robots. Works like [46], [45], and [47] propose multi-agent learning methods to approach the intersection management problem. Reference [38] proposes a learning based solution for optimizing robots' trajectories under bounded deviations of other robots from some nominal trajectories. Work [48] uses reinforcement learning to form platoons of vehicles and makes them pass through the intersection. In other multi-robot applications, there are works like [49] and [50] that learn path planning policies for navigation through narrow passages or hallways. Other works such as [51] present RL algorithms for general multi-robot trajectory coordination for purely first order kinematic robots. The core idea of our work has similarities to [52], [53], [54], and [55], which use RL methods to improve the efficiency of solving a parametrized combinatorial or mixed-integer optimization problem, for which the parameters are revealed online.

One of the main applications of our work is collision free warehouse automation. Prioritized/sequential planning seems to be a popular approach since it is both scalable and gives near-optimal solutions. Some works like [23], [56], [57], [58], and [59] assume the knowledge of some priorities over agents negotiating a path conflict (at an intersection), while [60] uses first come first serve policy for conflict resolution.

### B. DRAWBACKS OF CURRENT LITERATURE AND MOTIVATION

For computational scalability, several works decide the crossing/planning sequence (scheduling) among the robots and then get their trajectories by solving optimization problems in that order (prioritized/sequential planning). Scheduling is done using various techniques like auctions [32], job scheduling mechanisms [21] or some heuristics [14], [24], [25], [33]. The problem with this approach is that scheduling is dissociated from the trajectory generation and may lead to sub-optimal trajectories. Moreover, optimal policy for scheduling may be highly dependent on the type of intersection and other settings. Our past work [18] has a similar solution framework as in this paper. However, [18] contains no method or learning algorithm for obtaining a policy for determining crossing orders.

To the best of our knowledge, there are very few works like [28], [29], and [61] which make a systematic attempt at assigning priorities to the involved agents. Existing approaches to find a good set of priorities include iterative search algorithms [61] or solving an optimization problem [28], [29], which themselves are time consuming.

Although [39], [40], [41], [43], [44], [47] use RL to directly generate trajectories of the robots, the approach inherently cannot guarantee safety in a deterministic way. Further, these works only consider first order kinematic models of the robots, whereas we consider a double integrator model for the robots. Our proposed approach is very similar in spirit to [52], [53], [54], [55], which use learning methods to solve combinatorial optimization problems. However, these papers are for very different applications. Additionally, all these works provide algorithms that learn policies for a single agent rather than multiple agents.

### 1) MOTIVATION

Scalability, provable safety and efficiency are very important aspects for multi-robot systems like in autonomous warehouses since they directly affect the capital/operational cost. In light of the drawbacks of the existing literature, it is important to address all these three aspects in a combined manner. In this work, we address scalability, provable safety and efficiency together by proposing a way to combine learning and optimization methods. In particular, we use the optimization framework to ensure provable safety and use learning framework to improve scalability, while ensuring near-optimality.

### C. CONTRIBUTIONS

The following are the main contributions of this paper.

- We propose an algorithm for learning a policy for determining the order in which the robots cross an isolated intersection, given certain features of the traffic. The algorithm learns a policy that is shared among all the robots. We use this policy over a solution framework that combines learning and online optimization for unsignalized intersection management for a continual stream of robot traffic. The framework we use does sequential optimization of trajectories for each robot using the trajectories of other robots ahead of it in the sequence as constraints. This way, we ensure scalability as well as safety at all times, both during training and deployment. The framework implicitly handles continual stream of robot traffic.

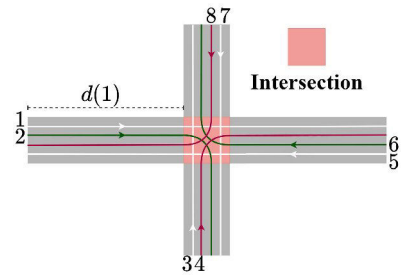

**FIGURE 1.** A schematic of an example intersection and the region of interest (RoI) with 8 lanes.

- Through extensive simulations, we establish that the proposed framework solves the intersection management problem in real-time and provides near-optimal solutions. In general, the performance of the trained policies is significantly superior compared to many heuristics from the literature. For example, in some scenarios, the learnt policies on average provide up to 150% improvement in terms of the considered objective function and up to 60% improvement in terms of average time to cross, when compared against the first come first serve heuristic. Also, in many empirical simulations, the trained RL policy outperforms all the compared heuristics by 40% in terms of the objective function and by 20% in terms of average time to cross weighted by robot priorities for a considerable range of arrival rates.
- We propose some adaptations to the underlying solution framework to address real-world implementation challenges like tracking errors and delays associated with communication and computation. The framework with these adaptations is deployed on physical robots to establish the real-time implementability of the framework i.e., communication and computations delays in the deployed framework are smaller than the spare-time available in the framework.

## II. UNSIGNALIZED INTERSECTION MANAGEMENT AND PROBLEM SETUP

In this section, we first present the unsignalized intersection management problem and discuss some challenges in solving it in real-time. Then, we pose a problem of learning computationally efficient and near-optimal policies that can be utilized for safe intersection management in real-time. We first describe the intersection geometry, robots and notation.

We consider an isolated *region of interest* (RoI), denoted as $\mathcal{R} \subset \mathbb{R}^2$, consisting of $M$ number of fixed lanes. Let $\mathcal{S}(l)$ for $l \in \{1, \ldots, M\}$ be the set of points in RoI $\mathcal{R}$ that form the $l^{\text{th}}$ lane. Without loss of generality, we assume that each lane $\mathcal{S}(l)$, for all $l \in \{1, \ldots, M\}$ is an open set. The *intersection* in the RoI $\mathcal{R}$ is

$$\mathcal{I} := \{z \in \mathcal{R} \mid z \in \mathcal{S}(l_1) \cap \mathcal{S}(l_2), \ l_1, l_2 \in \{1, \ldots, M\}, \ l_1 \neq l_2\},$$

i.e., the set of all points in the RoI that belong to at least two lanes. We assume that the intersection $\mathcal{I}$ is a connected set and that each lane leads to, goes through and leaves the intersection $\mathcal{I}$ only once.

Figure 1 shows one such example configuration, with the intersection $\mathcal{I}$ in the center. For a lane $l$, we denote its length of approach to the intersection as $d(l)$.

### A. DEFINITIONS OF ROBOT RELATED VARIABLES AND PARAMETERS

We assume that the robots travel only along the fixed lanes inside the RoI, i.e., they do not change lanes. We denote the set of robots under consideration by $V$. We denote the length of robot $i \in V$ with $L_i$. Further for a robot $i \in V$, we denote its lane number, time of arrival into the RoI, time of entry into and time of exit from the intersection by $q_i$, $t_i^{\mathcal{A}}$, $t_i^{\mathcal{E}}$ and $t_i^{\mathcal{X}}$, respectively. The time of arrival of a robot into the RoI is unknown before its arrival. We use $x_i(t)$, $v_i(t)$ and $u_i(t)$ to denote the longitudinal position (specifically, front end of the robot), longitudinal velocity and longitudinal acceleration of the robot, respectively, along its lane $q_i$ at time $t \geq 0$. We use $\dot{x}_i(t)$ and $\dot{v}_i(t)$ to denote the time derivative of $x_i(t)$ and $v_i(t)$ evaluated at time $t$, respectively. On each lane, we set up the coordinates such that for a robot $i \in V$ we have, $x_i(t_i^{\mathcal{A}}) = -d(q_i)$ and $x_i(t_i^{\mathcal{E}}) = 0$. To capture if two robots $i, j \in V$ are on lanes that intersect in the intersection area, we define a variable $c(i, j)$ as

$$c(i,j) = \begin{cases} 0 & \text{if } q_i = q_j \\ 1 & \text{if } q_i \neq q_j \text{ and } \mathcal{S}(q_i) \cap \mathcal{S}(q_j) = \emptyset \\ -1 & \text{if } q_i \neq q_j \text{ and } \mathcal{S}(q_i) \cap \mathcal{S}(q_j) \neq \emptyset, \end{cases} \quad (1)$$

where $\emptyset$ denotes the empty set. Hence, $c(i, j)$ is equal to 0, 1 and $-1$ if and only if the robots $i$ and $j$ are on the same lane, on different lanes that do not intersect and on different lanes that intersect, respectively. See Table 1 for a comprehensive list of important notations and their definitions used in this paper.

### B. CONSTRAINTS

We assume that the robots follow the double integrator dynamics with longitudinal acceleration as their input, i.e.,

$$\dot{x}_i(t) = v_i(t) \quad \dot{v}_i(t) = u_i(t), \ \forall i \in V, \ t \geq t_i^{\mathcal{A}}, \quad (2)$$

where, $\dot{x}_i(t)$ and $\dot{v}_i(t)$ are the time derivatives of position and velocity of robot $i$ at time $t$. $u_i(t)$ is the acceleration input to the robot $i$ at time $t$. $t_i^{\mathcal{A}}$ is the arrival time of robot $i$.

We assume that the robots' velocities and accelerations are bounded, i.e.,

$$v_i(t) \in [0, \bar{v}_i] \text{ and } u_i(t) \in [\underline{u}, \bar{u}], \quad \forall i \in V, \ \forall t \geq t_i^{\mathcal{A}} \quad (3)$$

where, $\underline{u} < 0$, $\bar{u} > 0$ and $\bar{v}_i > 0 \ \forall i \in V$.

To avoid collisions inside the intersection between robots on a pair of conflicting or incompatible lanes (pairs of lanes which intersect), we impose the constraint

$$t_i^{\mathcal{E}} \geq t_j^{\mathcal{X}} \text{ OR } t_j^{\mathcal{E}} \geq t_i^{\mathcal{X}}, \quad \forall i, j \in V, \text{ if } c(i,j) = -1, \quad (4)$$

where, $t_i^{\mathcal{E}}$ and $t_i^{\mathcal{X}}$ are time of intersection entry and exit of robot $i \in V$. Note that the constraint (4) is combinatorial.

Next, we formulate the rear-end safety constraint. In intersection management literature it is common to enforce that any two robots on a lane are separated by at least a fixed distance. However, note that in situations like warehouses, there is always a chance of robot or communication or coordination failure. Hence we propose a conservative rear-end safety constraint, which ensures the existence at all times, of a feasible input (acceleration) trajectory for a robot to come to a stop, avoiding collision, irrespective of the trajectory taken by the robot ahead. We formally describe this idea in the following definition and Lemma 1, which are adapted from [62]. Subsequently, we present the rear-end safety constraint.

*Definition 1 (Safe Following Distance [62]):* The maximum braking maneuver (MBM) of a robot is a control action that sets its acceleration to $\underline{u}$ until the robot comes to a stop, and its acceleration is set to 0 thereafter. For two robots $i, j \in V$ on the same lane with $i$ following $j$, i.e., $c(i, j) = 0$ and $t_i^{\mathcal{A}} > t_j^{\mathcal{A}}$, a quantity $\mathcal{D}(i, j, t)$ is a safe-following distance at time $t$ if $x_j(t) - x_i(t) \geq \mathcal{D}(i, j, t) \geq L_j$ and, if each of the two robots were to perform the MBM, then they would be safely separated, i.e. $x_j(\hat{t}) - x_i(\hat{t}) \geq L_j, \ \forall \hat{t} \in [t, \bar{t}]$, where $\bar{t} \geq t$ is the time when robot $j$ comes to a complete stop. •

*Lemma 1 (Minimum Safe Following distance [62]):* Let $i, j \in V$ be a pair of robots with $i$ following $j$ on the same lane, i.e., $c(i, j) = 0$ and $t_i^{\mathcal{A}} > t_j^{\mathcal{A}}$. Then, the continuous function

$$\mathcal{D}(i,j,t) = L_j + \max\left\{0, \frac{v_i^2(t) - v_j^2(t)}{-2\underline{u}}\right\} \quad (5)$$

provides a safe-following distance at time $t$ for the pair of robots $i$ and $j$. •

In light of this, we impose the rear-end safety constraints

$$x_j(t) - x_i(t) \geq L_j + \max\left\{0, \frac{v_i^2(t) - v_j^2(t)}{-2\underline{u}}\right\},$$

$$\forall t \geq t_i^{\mathcal{A}}, \forall i, j \in V, \text{ s.t. } c(i,j) = 0 \text{ and } t_i^{\mathcal{A}} > t_j^{\mathcal{A}}. \quad (6)$$

### C. CONTROL OBJECTIVE

Consider a set of robots $V$ that arrive into the RoI at different times during a time interval of interest. Since the robots arrive at different into the RoI at different times, we also call $V$ a stream of robots. For the robots in $V$, we want to minimize their cumulative time taken to cross the intersection weighted by their priorities, i.e.,

$$\min_{\{u_i : i \in V\}} \sum_{i \in V} r_i(t_i^{\mathcal{X}} - t_i^{\mathcal{A}})$$

$$\text{s.t.} \quad (2), (3), (6), \forall t \in [t_i^{\mathcal{A}}, t_i^{\mathcal{X}}], \ \forall i \in V, \ (4), \quad (7)$$

where $r_i > 0$ is a weight indicating the priority of robot $i$. Note that $t_i^{\mathcal{X}}$, for any $i \in V$, depends on the decision variable $u_i(.)$, which is the control input trajectory to robot $i \in V$. Hence, Problem (7) is a variable horizon optimal control problem, with each robot in $V$ having a different time horizon. This is particularly difficult to handle for a stream of robots that arrive at different times. Hence, we formulate the following proxy optimal control problem for intersection management for a stream of robots $V$.

**TABLE 1.** Table of some important notations.

| Notation | Definition |
|---|---|
| \multicolumn{2}{c}{RoI and intersection geometry related} | |
| $\mathcal{R}$ | Region of interest (RoI) |
| $M$ | Number of lanes |
| $\mathcal{S}(l)$ | Set of points in RoI, $\mathcal{R}$, that form the $l^{\text{th}}$ lane |
| $\mathcal{I}$ | Intersection |
| $d(l)$ | Length of approach of the $l^{\text{th}}$ lane |
| \multicolumn{2}{c}{Robot related} | |
| $q_i$ | Lane of robot $i$ |
| $x_i(t), v_i(t), u_i(t)$ | Position, velocity and acceleration (input), respectively, of robot $i$ along its lane at time $t$ |
| $\dot{x}_i(t), \dot{v}_i(t)$ | Time derivatives of position and velocity, respectively, of robot $i$ along its lane at time $t$ |
| $t_i^{\mathcal{A}}, t_i^{\mathcal{E}}, t_i^{\mathcal{X}}$ | Times of entry into RoI, entry into intersection and exit from intersection, respectively, of robot $i$ |
| $\bar{v}$ | Upper bound on robot velocities |
| $\underline{u}, \bar{u}$ | Lower and upper bounds, respectively, on the (input) acceleration for the robots |
| $c(i,j)$ | Indicator variable to represent lane compatibility of robots $i$ and $j$, see (1) |

$$\max_{\{u_i : i \in V\}} \sum_{i \in V} r_i \int_{t_i^{\mathcal{A}}}^{t_i^{\mathcal{A}} + T_h} v_i(t) \mathrm{d}t$$

$$\text{s.t.} \quad (2), (3), (6), \forall t \in [t_i^{\mathcal{A}}, t_i^{\mathcal{A}} + T_h], (4). \quad (8)$$

Here $T_h$ is a sufficiently long time horizon. We assume that the arrival time, $t_i^{\mathcal{A}}$, of a robot $i \in V$ is random and is unknown before its arrival. Notice that Problem (8) is a proxy for Problem (7) and we use the formulation (8) as it has the advantage of a fixed time horizon $T_h$ for each robot.

*Remark 1 (Challenges in Solving Problem (8) Directly):* There are several challenges in solving Problem (8). They mainly stem from the following factors.

(i) At any given time instant, the exact information about the future arrival of robots is not available.
(ii) Constraint (4) is combinatorial, which makes Problem (8) a mixed integer program. A naive optimization approach scales exponentially with the number of robots and lanes. This is problematic because intersection management problem is both time and safety critical.

In order to counter these challenges, we use a data-driven approach similar to the one in [18], wherein data obtained through simulations was used to manually tune the policy. In the current work, we propose algorithms that learn near-optimal policies to Problem (8).   •

### D. INFORMAL PROBLEM STATEMENT

In this work, we seek to develop algorithms that learn near-optimal policies for Problem (8). While we allow the learning to be centralized, implementation of learnt policies should be distributed, i.e., the policy for each robot should depend on only the data that can be easily obtained using robot-to-robot and robot-to-infrastructure communication. The proposed solution should be applicable to a continual stream of robots arriving randomly into the RoI. Note that *any feasible solution to Problem (8) implicitly guarantees safety.*

### III. SOLUTION FRAMEWORK TO ADDRESS RANDOM ARRIVAL TIMES

In this section, we present an overall framework/algorithm for solving the intersection management problem. The broad solution framework is a modified version of the one in [18]. We recap the main aspects of this framework.

As the arrival times of robots are unknown beforehand and coordination between robots requires planning for groups of robots at a time, we split the trajectory of each robot into two phases - *provisional phase* and *coordination phase*. Provisional phase of a robot begins when it arrives at the RoI and ends when its coordination phase begins. A coordination phase algorithm runs every $T_c$ seconds and assigns safe and efficient trajectories for crossing the intersection to the robots in their provisional phase.

Before we discuss the specifics of this framework, we introduce some notation. For a robot $i \in V$, $t_i^{\mathcal{C}} \geq t_i^{\mathcal{A}}$ represents the time at which its coordinated phase starts, and hence $t_i^{\mathcal{C}} = kT_c$ for some $k \in \{1, 2, \ldots\}$. We let $V(k) := \{i \in V : t_i^{\mathcal{A}} < kT_c\}$ be the set of robots which entered the RoI before $kT_c$, $V_s(k) := \{i \in V(k) : t_i^{\mathcal{C}} < kT_c\}$ be the robots that entered coordinated phase before $kT_c$, $V_p(k) := V(k) \setminus V_s(k)$, i.e., the set of robots that need coordinated phase trajectories at $kT_c$.

## A. PROVISIONAL PHASE

Consider a robot $i \in V_p(k)$ ($k \in \mathbb{N}$). To ensure that the robot does not enter the intersection before it enters the coordinated phase, we impose the constraint

$$v_i(t) \leq \sqrt{2\underline{u}x_i(t)}. \tag{9}$$

We let the robot's acceleration input for the provisional phase trajectory in the time interval $[\max\{t_i^{\mathcal{A}}, (k-1)T_c\}, kT_c]$ be an optimal solution of the following problem.

$$\max_{u_i(.)} \int_{\max\{t_i^{\mathcal{A}}, (k-1)T_c\}}^{kT_c} v_i(t) \mathrm{d}t$$

s.t. (2), (3), (6), (9), $\forall t \in [\max\{t_i^{\mathcal{A}}, (k-1)T_c\}, kT_c]$. (10)

## B. COORDINATED PHASE

At the time $kT_c$, for each $k \in \mathbb{N}$, some or all the robots in $V_p(k)$ are assigned their coordinated phase trajectories, which they start executing immediately. Ideally, we would like the coordinated phase control input trajectories for the robots in $V_p(k)$ to be an optimal solution of Problem (11), which we call as the *combined optimization* problem.

$$\max_{\{u_i : i \in V_p(k)\}} \sum_{i \in V_p(k)} r_i \int_{kT_c}^{kT_c+T_h} v_i(t) \mathrm{d}t$$

s.t. (2), (3), (6), $\forall t \in [kT_c, kT_c + T_h]$, (4). (11)

Note that the robots in $V_s(k)$ would also appear in the constraints of Problem (11). Intersection safety constraint in Problem (11) is still combinatorial. Solving Problem (11) inherently involves picking a feasible solution with highest objective value among all the feasible crossing orders. As a result, this formulation scales exponentially with the number of robots and is not suitable for real-time implementation. We illustrate this exponential scaling of computation times through simulations in Figure 5.

Algorithm 1 presents the overall framework for coordinating the stream of robots. It describes when provisional and coordinated phases are run and on what robots. Algorithm 1 perpetually checks for new robots entering the RoI and assigns them provisional phase trajectories as and when they arrive. Periodically, with period $T_c$, the algorithm computes coordinated phase trajectories for all the robots in their provisional phase. Among these robots, those that can cross the intersection with the computed trajectory start executing it and hence enter their coordinated phase. On the other hand, the robots that cannot cross the intersection with the newly computed trajectories, continue in provisional phase with updated provisional phase trajectories, computed as per (10).

*Remark 2 (Possibility of Multiple Provisional Phases for a Robot):* In busy intersections, there might be cases where the solution to Problem (11) gives a trajectory for a robot

---

**Algorithm 1** `overall_algorithm`
**Inputs**: Coordinated Phase Time Period $T_c$.

1   $k = 1$, $V_p(1) = \emptyset$, $V_s(1) = \emptyset$
2   **while** *True* **do**
3     **if** `robot enters RoI` **then**
4       Run `provisional_phase` for robot
5       Add robot to $V_p(k)$
6     **end**
7     **if** `time` $= kT_c$ **then**
8       Compute `coordinated_phase` on $V_p(k)$
9       Set $V_{nc}(k)$ as the set of robots among $V_p(k)$ for which trajectories computed for the coordinated phase do not let them cross the intersection
10       Set $V_s(k+1) = V_s(k) \cup V_p(k) \setminus V_{nc}(k)$ and $V_p(k+1) = \emptyset$
11       **for** $i \in V_{nc}(k)$ **do**
12         Run `provisional_phase` for $i$
13         Add $i$ to $V_p(k+1)$
14       **end**
15       Set $k = k + 1$
16     **end**
17 **end**

---

$i \in V_p(k)$ such that the robot does not exit the intersection by $kT_c + T_h$. This can potentially cause infeasibility of Problem (11) in subsequent coordinated phases. In order to avoid such a situation, such affected robots (robot $i$ and its successors in the crossing order) go through the provisional phase again for the interval $[kT_c, (k+1)T_c]$. This is the reason why the time horizon for robot $i \in V_p(k)$ is $[\max\{t_i^{\mathcal{A}}, (k-1)T_c\}, kT_c]$ in Problem (10). We emphasize the difference between a robot going through coordinated phase computation and going through coordinated phase itself (using the computed coordinated phase trajectories). Specifically, in Algorithm 1, the robots in $V_p(k)$ are going through coordinated phase computation and at the end of $k^{\text{th}}$ iteration in Algorithm 1, only the robots in $V_s(k+1) \setminus V_s(k)$ start their coordinated phase. Robots in $V_{nc}(k)$ will go through a provisional phase again.   •

## C. SEQUENTIAL OPTIMIZATION FOR COORDINATED PHASE

To address non-scalability of combined optimization for obtaining coordinated phase trajectories, we present a modified version of sequential optimization from [18] in Algorithm 2. The algorithm takes as input a set of quantities called *precedence indices* ($p_i \in \mathbb{R}$, $\forall i \in V_p(k)$), which determine the crossing order. Each robot in $V_p(k)$ obtains its coordinated phase trajectory sequentially, as per the crossing order, by solving the optimization problem in (12).

In Algorithm 2, at the beginning of each iteration of the while loop, $V_Q$ is the subset of robots in $V_p(k)$ that do not have a coordinated phase trajectory yet. In Step 3, $\mathcal{F}$ is the

**Algorithm 2** `sequential_optimization`
**Inputs**: $V_p(k)$, $V_s(k)$, $p_i \; \forall i \in V_p(k)$

1  $V_Q = V_p(k)$
2  **if** $|V_Q| > 0$ **then**
3      $\mathcal{F} \leftarrow \{i \in V_Q : x_i \geq x_j, \; \forall j \in V_Q \text{ s.t. } q_i = q_j\}$
4      $i^* \leftarrow \underset{j \in \mathcal{F}}{\arg\max} \; \{p_j\}$
5      $V_s^{i^*} \leftarrow V_s(k) \cup (V_p(k) \setminus V_Q)$
6      $u_{i^*}(.) \leftarrow$ `Solve (12) for robot` $i^*$
7      **if** $i^*$ `crosses the intersection by time` $kT_c + T_h$ `with` $u_{i^*}(.)$, **then**
8        $t_{i^*}^{\mathcal{C}} \leftarrow kT_c$
9        $i^*$ starts executing $u_i^*$ at $t_{i^*}^{\mathcal{C}}$
10       `Remove` $i^*$ `from` $V_Q$
11       `Go back to line 2`
12     **end**
13 **end**

set of robots in $V_Q$ that are nearest to the intersection in their respective lanes. In Step 4, we obtain the robot $i^*$ in $\mathcal{F}$ with the largest precedence index $p_{i^*}$, after breaking ties arbitrarily. In Step 5, $V_s^{i^*}$ is the set of all robots that were assigned a coordinated phase trajectory before the robot $i^*$. If $u_{i^*}^*(.)$, computed in Step 6, enables the robot to cross the intersection before $kT_c + T_h$, then the robot $i^*$ starts executing $u_{i^*}^*(.)$ as its coordinated phase control trajectory, starting at $t_{i^*}^{\mathcal{C}} = kT_c$. Then, $i^*$ is removed from $V_Q$ and the loop continues. On the other hand, if $i^*$ cannot cross the intersection with the control trajectory $u_{i^*}(.)$ then we break out of the loop and $i^*$ and the rest of the robots in $V_Q$ go through another provisional phase, as described in Remark 2.

In Step 6 of Algorithm 2, $u_{i^*}(.)$ is obtained by solving the following optimization problem.

$$J_{i^*}^* = \max_{u_{i^*}(.)} \int_{kT_c}^{kT_c+T_h} v_{i^*}(t) dt$$

s.t. (2), (3), (6), $\forall t \in [kT_c, kT_c + T_h]$, with $i = i^*$,
and $t_{i^*}^{\mathcal{E}} \geq \tau_{i^*}$, (12)

where $\tau_{i^*}$ is the *minimum wait time* of $i^* \in V_p(k)$, given by,

$$\tau_{i^*} = \max\{t_m^{\mathcal{X}} : m \in V_s^{i^*} \text{ s.t. } q_{i^*} \text{ and } q_m \text{ intersect}\}.$$

*Remark 3 (Features of Algorithm 2):* Algorithm 2 has several good features, which help in achieving the design goals of scalability and real-time implementation. In a variety of simulations, we have consistently observed that the computation for Algorithm 2 scales linearly with the number of robots. We illustrate this feature in Figure 5. Algorithm 2 is an efficient version of the DD-SWA algorithm proposed in [18], as the precedence indices need not be recomputed after the trajectory optimization for each robot. The solution framework, including Algorithm 2, can be implemented in a distributed manner. Reader may refer to relevant discussions in [18].
    ●

The solution framework is recursively feasible under the assumption that every robot enters the RoI satisfying constraints (3), (6) and (9). Theorem 1 formalizes the result guaranteeing the safety of the system for all time. This result is very similar to the one in [18].

*Theorem 1 (System Wide Recursive Safety [18]):* If every robot $i \in V$ satisfies the rear-end safety constraint (6) at the time of its arrival, $t_i^{\mathcal{A}}$, and its initial velocity is such that $v_i(t_i^{\mathcal{A}}) \leq \min(\bar{v}_i, \sqrt{2\underline{u}x_i(t_i^{\mathcal{A}})})$, feasibility of problems (10), (11) and (12) is guaranteed. Consequently, safety of all the robots is also guaranteed for all time under Algorithm 1.

*Proof:* Recall that each robot $i \in V$ is allowed to decelerate i.e., $\underline{u}_i < 0$. Hence, due to the assumption that robot's entry velocity, $v_i(t_i^{\mathcal{A}})$ is less than the allowed upper-bound and satisfies the rear-end safety constraint (6), infeasibility cannot occur due to the violation (6). Further, it is also assumed that $v_i(t_i^{\mathcal{A}}) \leq \sqrt{2\underline{u}x_i(t_i^{\mathcal{A}})}$, ensuring the existence of a feasible control trajectory so that the robot comes to a stop before entering the intersection, making Problem (10) feasible.

Notice that if Problem (10) is feasible, the trajectory of provisional phase guarantees for robot $i$ that when it goes through the coordinated phase computations, say at $t_i^{\mathcal{C}}$, the intersection safety constraint (4) is satisfied at $t_i^{\mathcal{C}}$. This inherently ensures that $t_i^{\mathcal{E}} \geq \tau_i$ is feasible (recall $\tau_i$ is the minimum wait time for $i$). This guarantees feasibility of problems (11) and (12).   □

For the solution framework to be complete, we still need to specify how the precedence indices $p_i$ are to be chosen. In the next section, we propose an algorithm for learning a policy that gives out "near-optimal" precedence indices, given some information about the traffic state.

## IV. LEARNING A POLICY THAT GIVES NEAR-OPTIMAL CROSSING ORDERS

Recall that at $kT_c$, the beginning of the $k^{\text{th}}$ coordinated phase, $V_p(k)$ is the set of robots that need coordinated phase trajectories and $V_s(k)$ is the set of robots that are already executing their coordinated phase trajectories. Notice that Algorithm 2 takes as input the precedence indices of the robots in $V_p(k)$, using which it sequentially optimizes the coordinated phase trajectories of the robots. In this section, we are interested in obtaining a policy that determines the precedence indices of the robots in $V_p(k)$, given the traffic state, so that Algorithm 2 provides optimal or at least near-optimal solutions to the combined optimization problem (11) and more generally to the original optimization problem (8).

In particular, we propose a centralized algorithm for learning a policy, which

(i) can be implemented online in real-time.
(ii) is shared, i.e., the same policy is used by all the robots.

(iii) is distributed, i.e., a policy to which the inputs are information available to a robot locally or through communication with its neighbouring robots.
(iv) can be implemented on an arbitrary number of robots.

We denote the shared policy by the function $g(.)$. For robot $i$, the input to the policy is the *feature vector*, $f_i$, that captures the state of the traffic relevant to robot $i$. Then, the precedence indices are

$$p_i = g(f_i), \quad \forall i \in V_p(k). \tag{13}$$

There may be different number of robots in $V_p(k)$ for different $k \in \mathbb{N}$. However, given the dimensions of the RoI and lengths of the robots we can determine $N_r$, an upper bound on the number of robots that could ever be in $V_p(k)$. Then, we pad the set of robots in $V_p(k)$ with $N_r - |V_p(k)|$ number of *pseudo robots* so that the state and action space dimensions remain constant for each $k \in \mathbb{N}$. Pseudo robots are virtual robots with features chosen such that they do not affect the feasibility and optimality of the crossing order and trajectories of the real robots. For e.g., the position of a pseudo robot can be picked far away from entry to RoI.

### A. MARKOV DECISION PROCESS FORMULATION

Let $\tilde{V}(k)$ be the union of $V_p(k)$ and the set of pseudo robots, so that $|\tilde{V}(k)| = N_r$. Then, we consider the Markov decision process (MDP) with the state, action, reward and the next state at the $k^{\text{th}}$ iteration defined as follows.

(i) The state, $s$, is the vector formed by concatenated feature vectors of all the robots in $\tilde{V}(k)$.
(ii) The action, $a$, is the vector of precedence indices of all the robots in $\tilde{V}(k)$ (by extension, the crossing order).
(iii) The reward, $R$, is as given in (14).
(iv) The next state, $s'$, is the concatenated feature vectors of the robots in $\tilde{V}(k+1)$.

The state space can be described as the set of all concatenated feature vectors (including those of an appropriate number of pseudo robots). Similarly, the action space is the set of all precedence index vectors of all the robots (including pseudo robots), i.e., $\mathbb{R}^{N_r}$.

Given a state at $k^{\text{th}}$ coordination phase computation, the RL agent decides an action as precedence indices indicating the crossing order. This is fed as an input to the sequential optimization (Algorithm 2), which outputs coordinated phase trajectories for the robots. These trajectories along with the trajectories of the robots in their provisional phase during $[kT_c, (k+1)T_c)$ determine the state $s'$ at the next, $(k+1)^{\text{th}}$, coordinated phase computation time. This way, the provisional phase and the sequential optimization algorithms act as the environment to the RL agent.

*Remark 4 (Reward Function):* Consider the $k^{\text{th}}$ coordinated phase. Given the action $a$ (precedence indices), let

$$V_c(k) := \{i \in V_p(k) : t_i^{\mathcal{C}} = kT_c\}, \ V_{nc}(k) := V_p(k) \setminus V_c(k),$$

where $t_i^{\mathcal{C}}$ is the time at which robot $i$ begins its coordinated phase, according to Algorithm 2. Thus, $V_c(k)$ is the set of robots that begin their coordinated phase at $kT_c$, while the robots in $V_{nc}(k)$ undergo another provisional phase. Then the reward function is

$$R = \sum_{i \in V_c(k)} r_i \frac{x_i(t_i^{\mathcal{A}} + T_r) - x_i(t_i^{\mathcal{A}})}{|V_p(k)|} - \sum_{j \in V_{nc}(k)} \bar{r} \frac{(d(q_j) - x_j(kT_c))^2}{|V_p(k)|}, \tag{14}$$

where $T_r$ is a suitable time horizon for the computation of the reward and $\bar{r} = \max_{i \in V} r_i$ is the maximum priority among all the the considered robots. The first term is the weighted sum of distances covered by the robots in $V_c(k)$ during a time interval of length $T_r$ since their arrival into RoI. As Algorithm 2 does not provide a coordinated phase trajectory for the robots in $V_{nc}(k)$, for each robot in $V_{nc}(k)$, we have a penalty term proportional to the square of the distance covered by the robot from its time of arrival up to $kT_c$. This penalty helps in learning a near-optimal precedence index policy (13). We observed that penalizing all the robots with the equal weight of $\bar{r}$ (in contrast to weighing penalty terms by individual priorities) leads to better policies. This may be due to the reason that a low priority robot crossing quickly makes way for a higher priority robot which will arrive in the future to travel quickly. •

#### 1) MULTI-AGENT JOINT-ACTION DDPG (MAJA-DDPG)

We use a modified version of the centralized multi-agent Deep Deterministic Policy Gradient (MA-DDPG [63]) algorithm to learn the precedence index policy (13). In this framework, the shared policy $g(.)$ is encoded by an actor neural network. In addition, there are neural networks encoding a target actor, a critic network encoding the estimated action value function $Q(.)$ and a target critic network $\hat{Q}(.)$. Let $\theta, \hat{\theta}, \phi$ and $\hat{\phi}$ be the parameters of the actor, target actor, critic and target critic networks, respectively. We store the (joint-state, joint-action, reward and joint-next-state) tuples in the replay buffer and use samples from this replay buffer to update the central-critic and shared actor networks in each learning iteration.

Suppose that $(s_m, a_m, R_m, s'_m)$ is the $m^{\text{th}}$ sampled tuple out of $N$ samples. The critic is updated to minimize the loss function $E$, given in (15). The actor is updated according to "gradient" ascent of the sampled gradient of estimated return $J$ with respect to actor parameters $\theta$, as in (16).

$$E = \frac{1}{N} \sum_m (R_m + \gamma \hat{Q}(s'_m, \hat{G}(s'_m)) - Q(s_m, a_m))^2. \tag{15}$$

$$\nabla_\theta J = \frac{1}{N} \sum_m \nabla_a Q(s, a)|_{s=s_m, a=G(s_m)} \nabla_\theta G(s)|_{s=s_m}. \tag{16}$$

Here $G(s)$ and $\hat{G}(s)$ represent the joint action (concatenated precedence indices of all robots in $V_p(k)$) and joint target action of all the robots in $V_p(k)$, respectively. Rest of the updates for target networks are similar to what is followed in the DDPG algorithm in [64].

### 2) ONLINE AND OFFLINE LEARNING APPROACHES

Algorithm 3 is the online approach to the proposed learning algorithm. We generate streams of robots using a poisson process for determining the arrival times of robots and choosing their initial velocities randomly. All these robots go through provisional and coordinated phases as described in Section III, where Algorithm 2 is used for obtaining coordinated phase trajectories. Since in Algorithm 2, we only care about the relative order of the precedence indices, we have a softmax layer (only during training) as the last layer in the RL actor network and use these values as the precedence indices of corresponding robots. We observed that having this layer leads to quicker learning since it limits the possibility of different actions leading to same crossing order. We add the corresponding state, action, reward and next state tuple to the replay buffer and update the critic, actor and the target networks as described above. Note that because of the softmax layer, the implementation of this algorithm cannot be done in a truly distributed manner. We use a modified Ornstein-Uhlenbeck process (with decaying variance in added Gaussian noise) as the exploration noise. Algorithm 3 can be run in an online loop continuously, gathering data and learning from it. A schematic flow of Algorithm 3 is presented in Figure 2.

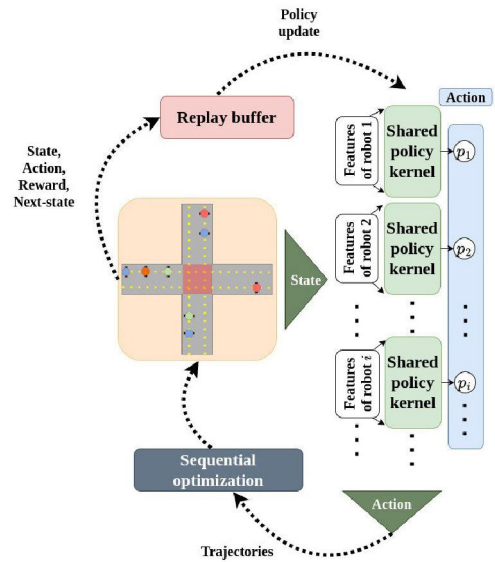

**FIGURE 2.** A schematic of online learning algorithm to train the shared policy.

---

**Algorithm 3** online_learning
**Inputs**: $V_p(k)$, $V_s(k)$, Shared-Policy $g$, Replay-Buffer

1  $\tilde{V}(k) \leftarrow V_p(k) \cup$ pseudo robots
2  s $\leftarrow$ concatenated feature vectors of robots in $\tilde{V}(k)$
3  $P \leftarrow \left(g(f_j)\right)_{j=1}^{j=|\tilde{V}|}$
4  a $\leftarrow$ softmax(P) + exploration_noise
5  sequential_optimization $\left(V_p(k), V_s(k), \{a_j : j \in V_p(k)\}\right)$
6  Compute reward $R$ according to (14)
7  $s' \leftarrow$ concatenated feature vectors of robots in $\tilde{V}(k+1)$
8  Add $(s, a, R, s')$ tuple to replay-buffer
9  Update $g$ and $Q$ as in [64] using (15) and (16)
10 Update targets $\hat{g}$ and $\hat{Q}$ using a polyak factor as in [64]
11 Return replay-buffer, $g$

---

Similar to DDPG and MADDPG, the proposed algorithm can also be used for offline learning. Given a data-rich replay buffer, the mini-batches can be sampled from this replay buffer and the RL agent's actor and critic networks can be updated iteratively. We propose an offline learning approach which involves constructing multiple *individual replay buffers*, one for each average arrival rate of robots using the steps as indicated in the online-approach and then merging them to form a new *merged replay buffer*. This merged replay buffer is then used to train a common policy, which can then be deployed for a range of average arrival rates. We refer to this as the Collect-Merge-Learn (CML) approach. A schematic of this approach is presented in the Figure 4. In all our simulations and experimental results, we use the shared policy kernel, $g(.)$, represented by a very simple neural network of the architecture whose schematic is presented in Figure 3 (see Remark 7 for details).

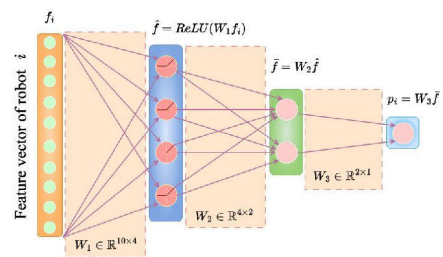

**FIGURE 3.** A schematic of the shared policy neural network.

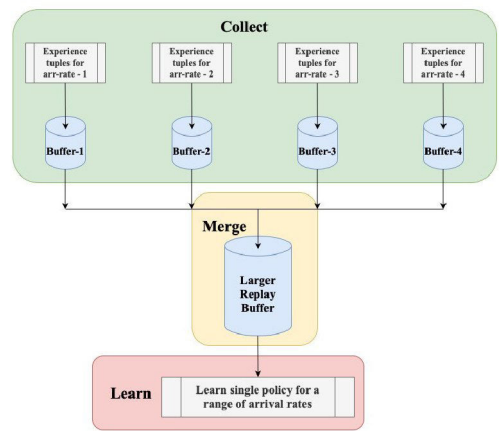

**FIGURE 4.** A schematic of offline learning framework using Collect-Merge-Learn (CML) approach.

*Remark 5 (Differences Between DDPG [64], MAD-DPG [63] and MAJA-DDPG):* While DDPG, MADDPG and MAJA-DDPG are all centralized learning algorithms, DDPG learns a policy for a single agent. Both MADDPG and MAJA-DDPG learn a shared policy for multiple agents. The main difference between MADDPG and our proposed MAJA-DDPG is in the computation of the sampled gradient of $J$ in (16). MAJA-DDPG uses the gradient of the critic action value function $Q$ with respect to the shared actor parameters $\theta$ through the joint action (concatenated actions of all the agents), whereas MADDPG estimates the gradient of $Q$ with respect to the shared actor parameters $\theta$ through the action of a randomly selected agent in the corresponding computation. •

*Remark 6 (List of Features of a Robot):* We consider the following quantities as features of a robot $i \in V_p(k)$: its distance from the entry to RoI ($x_i(kT_c) - d(q_i)$), its current velocity ($v_i(kT_c)$), its priority ($r_i$), its lane identifier ($q_i$), upper-bound on its velocity ($\bar{v}_i$), upper-bound on acceleration ($\bar{u}$), time since its arrival into RoI ($kT_c - t_i^A$), its estimated minimum wait time ($\bar{\tau}_i = \max\{t_j^{\mathcal{X}} : j \in V_s(k)\}$), the number of robots following on its lane, and the average distance between the robots following it on its lane. As can be seen, most of these features can be directly measured or computed by robot $i$, whereas the others can be obtained by communicating with neighbouring robots. •

*Remark 7 (Structure of the Shared Policy):* We use a neural network to capture the shared policy, $g(.)$, which gives the precedence index $p_i$ of a robot $i \in V$. Figure 3 presents a schematic of this neural network. The input to this network is the features, $f_i$ of the robot $i \in V$, ($f_i \in \mathbb{R}^{10}$ since we use 10 features for a robot as listed in Remark 6). This input is processed through two hidden layers. The first hidden layer has 4 units with ReLU activation and the other hidden layer has 2 units with linear activation. A linear combination of the outputs at the last hidden layer is taken as the output of the policy (the precedence index). •

*Remark 8 (Centralized and Distributed Implementations of Trained Policy):* Notice that once a policy is trained, it can be implemented either in a centralized or distributed manner, with the help of robot to infrastructure (R2I) and robot to robot (R2R) communication. For centralized implementation, a central intersection manager collects and maintains the up-to-date information about the state of all the robots in the ROI. The provisional phase for new robots and coordinated phases for existing robots can be computed in a centralized way (following Algorithm 2) and faithfully relayed to the vehicles through the R2I communication. For a distributed implementation, notice that all the features of a robot can be constructed using local information and can be acquired by a chain of R2R and R2I links. Also, the softmax activation in computing precedence indices can be ignored. Moreover, trajectories for a robot in the coordinated phase can also be computed locally on that robot, after all the required information for that robot-specific optimization problem being communicated using R2R and R2I communication.

We assume that all communications are instantaneous and error-free. Hence discussions on required bandwidth, specific communication protocols to be used etc. are outside the scope of this work. A more detailed discussion of a distributed implementation can also be found in [18]. •

## V. SIMULATION RESULTS

In this section, we present simulations comparing the proposed learning based sequential optimization algorithm against combined optimization and some other policies from literature.

The source code for the simulations is available at the link https://github.com/Control-Network-Systems-Group-IISc/IntMan-SeqOpt-Learn.

### A. SIMULATION SETUP
#### 1) SIMULATION PARAMETERS

We consider a warehouse scenario with 8 lanes meeting in the intersection from 4 directions. Each lane has an approach length of $7m$, i.e., $d(l) = 7m$, $\forall l \in \{1, 2, \ldots, 8\}$ in Figure 1. Each lane is $0.7m$ wide, making the intersection a square of side $2.8m$. For each robot $i$, its length $L_i$ is $0.75m$ and upper and lower bounds on its acceleration are $2m/s^2$ and $-2m/s^2$, respectively. The initial velocity of robot $i$ is sampled from a uniform distribution on $[0, \bar{v}_i]$. We carried out simulations for streams of randomly arriving robots with various *average arrival rates*. In particular, streams were generated by choosing tentative arrival times of robots into the lanes according to a poisson process, with a specified average arrival rate. The actual arrival time of each robot is delayed till the rear-end-safety condition (6) is satisfied. We set the coordinated phase computation interval to $6s$, i.e., $T_c = 6s$.

We say that a simulation has homogeneous traffic if all the lanes have the same average arrival rate and heterogeneous traffic otherwise. If the (lane dependent) average arrival rate remains same throughout the duration of simulation, we say that the simulation has static traffic and time-varying otherwise. For the simulations having static heterogeneous traffic we choose the average arrival rates on different lanes to be 0.13, 0.18, 0.08, 0.15, 0.19, 0.09, 0.05 and 0.16 robots/lane/s on lanes 1, 2, 3, 4, 5, 6, 7 and 8 respectively. We further differentiate how the average arrival rates vary with time by saying that the simulation has random-time-varying traffic or burst-mode-time-varying traffic. In the case of random-time-varying traffic we choose to sample the average arrival rate on each lane uniformly from {0.05, 0.06, 0.07, 0.08, 0.09, 0.1, 0.11, 0.12, 0.13, 0.14, 0.15} every 100s. In case of burst-mode-time-varying traffic, every 30s, we choose to set average arrival rate to be 0.15 robots/lane/s for the first 10s and 0.05 robots/lane/s for the remaining 20s.

A simulation is said to have homogeneous parameters if all the robots involved in the simulation share the same priority and upper bound on velocity i.e., $r_i = 1$ and $\bar{v}_i = 1.5$m/s $\forall i \in V$. On the other hand, the simulation is said to have heterogeneous parameters if the priorities and velocity upper bounds are different for different robots. Specifically, we choose to set the robot priorities randomly by sampling from the set $\{1, 2, 4, 5\}$ with probability $0.5, 0.3, 0.15$ and $0.05$ respectively at the time of its entry into RoI. We choose to set lane dependent velocity upper bounds where robots on lanes 1, 4, 5 and 8 have velocity upper bound as 1.5m/s and those on lanes 2, 3, 6 and 7 have velocity upper bound as 1m/s.

Different simulation settings we use to study the performance of the proposed approach is presented in Table 3. Each cell that shows homogeneous and static traffic, also shows the set of arrival rates in robots/lane/s for which the corresponding training or testing simulations were done.

### 2) CML TRAINING

For training the policies, we collect 10 individual replay buffers (one for each average arrival rate for cases with training on homogeneous traffic setting and 10 of the same average arrival rate setting for cases with training on heterogeneous traffic setting), each containing data of 5000 coordinated phases collected from random streams (evolving according to Algorithm 3). These individual replay buffers are merged to form a single merged replay buffer. This merged replay buffer is then used to train, using the CML approach, a set of 10 policies (with different network initializations). We use a discount factor of 0.99 for all training simulations.

### B. SUB-OPTIMALITY AND COMPUTATION TIMES OF SEQUENTIAL OPTIMIZATION METHOD

In this subsection, we first demonstrate that, in general, sequential optimization can give near-optimal solutions. For this, we compare the performance of combined optimization against sequential optimization for all possible crossing orders and choosing the one with the best crossing order by exhaustive search, which we call as BESTSEQ. We also illustrate that there is a tremendous computational advantage of sequential optimization, with a CML policy determining the precedence indices over combined optimization and BESTSEQ.

For these comparisons, we collected coordinated phase problem instances ($V_p(.)$ and $V_s(.)$) from 3 streams of 500s each, with an average arrival rate of 0.08 robots/lane/s with BESTSEQ (homogeneous, static traffic and heterogeneous parameter setting with $T_h = 30s$). For each coordinated phase instance, we also computed the optimum solution using combined optimization.

For a given coordinated phase problem instance, let $C_{CO}$ and $C_{BS}$ denote the objective values of Problem (11) obtained using combined optimization and BESTSEQ respectively. Since combined optimization is computationally expensive, we consider only those coordinated phases with $|V_p(.)| \leq 6$, making a total of 217 coordinated phase instances. Table 2 presents the number of problem instances (No. of inst.),

**TABLE 2.** Sub-Optimality of BESTSEQ.

| $\lvert V_p(.) \rvert$ | | 1 | 2 | 3 | 4 | 5 | 6 |
|---|---|---|---|---|---|---|---|
| No. of inst. | | 21 | 42 | 34 | 49 | 55 | 16 |
| opt_gap (%) | avg | 0 | 0.74 | 1.19 | 1.44 | 2.04 | 2.05 |
| | 90[th]p | 0 | 2.07 | 2.86 | 2.90 | 3.69 | 3.29 |

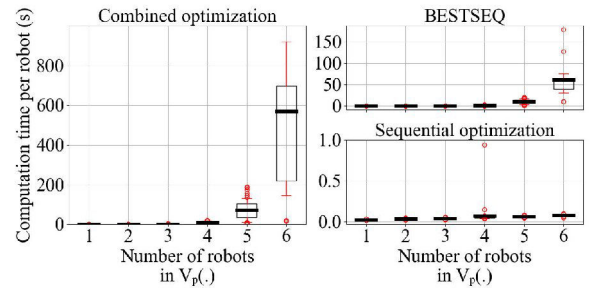

**FIGURE 5.** Computation time per-robot for combined optimization, BESTSEQ and CML trained sequential optimization.

average (avg) and 90[th] percentile (90[th]p) of optimality gaps,

$$\text{opt\_gap} := \frac{C_{CO} - C_{BS}}{C_{CO}}\%$$

We see that, on average, the sub-optimality of BESTSEQ is within an acceptable range ($\sim 2\%$).

For the same set of problem instances, computation time per robot for combined optimization, BESTSEQ and sequential optimization, with a CML trained policy, are compared in Figure 5. In this figure, bold black lines represent the mean, boxes represent the range of values between the first and third quartile, the whiskers represent the 10[th] and 90[th] percentile and red circles are outliers. We see that, with some sub-optimality, BESTSEQ incurs far less computation times compared to combined optimization. The computation time per robot for sequential optimization, with a CML trained policy, is essentially constant with the number of robots and several orders of magnitude lower than that of BESTSEQ and combined optimization - specially for higher number of robots. Thus, our proposed framework is far more suitable for real-time implementation.

### C. COMPARISON OF CML TRAINED POLICIES AGAINST OTHER POLICIES

Next, we compare the performance of CML trained policies against some heuristics from literature. Since the computation time for combined optimization is prohibitive except for very low traffic arrival rates, we skip it from these comparisons. Though BESTSEQ is not as computationally intensive as combined optimization, it too is prohibitively costly and we report data from BESTSEQ only for low traffic conditions. Given its small optimality gap with respect to combined optimization, it also serves as a reasonable benchmark where its computation time is manageable.

**TABLE 3.** Traffic parameters and other parameters used in different simulations. Acronyms: Hm. - Homogeneous, Ht. - Heterogeneous, Param. - Parameters. Note: For Sim-1 to Sim-5, we train separate CML policies for each simulation. But for testing Sim-6 to Sim-9 we use the policies trained in Sim-2.

|  | Training | | | | Testing | | |
| --- | --- | --- | --- | --- | --- | --- | --- |
|  | Traffic | Param. | $T_h$ (s) | $T_r$ (s) | Traffic | Param. | $T_h$ (s) |
| Sim-1 | Hm., Static {0.01, 0.02, ..., 0.1} | Ht. | 30 | 20 | Hm., Static {0.01, 0.02, ..., 0.1} | Ht. | 30 |
| Sim-2 | Hm., Static {0.11, 0.12, ..., 0.2} |  | 60 | 30 | Hm., Static {0.11, 0.12, ..., 0.2} |  | 60 |
| Sim-3 | Hm., Static {0.01, 0.02, ..., 0.1} | Hm. | 30 | 20 | Hm., Static {0.01, 0.02, ..., 0.1} | Hm. | 30 |
| Sim-4 | Hm., Static {0.11, 0.12, ..., 0.2} |  | 60 | 30 | Hm., Static {0.11, 0.12, ..., 0.2} |  | 60 |
| Sim-5 | Ht., Static | Ht. | 60 | 30 | Ht., Static | Ht. | 60 |
| Sim-6 | Hm., Static {0.11, 0.12, ..., 0.2} | Ht. | 60 | 30 | Hm., Static {0.01, 0.02, ..., 0.1} | Ht. | 30 |
| Sim-7 |  |  |  |  | Hm., Static {0.125, 0.175, 0.21, 0.22, ..., 0.3} |  | 60 |
| Sim-8 |  |  |  |  | Hm., burst-mode time-varying |  | 30 |
| Sim-9 |  |  |  |  | Ht., random time-varying |  |  |

In the overall solution framework described in Section III, we compare the performance of several policies for determining the crossing order in sequential optimization, Algorithm 2. We compare the policies generated by CML against the following policies for various situations.

(i) BESTSEQ: Sequential optimization for the best crossing order, which is determined with an exhaustive search.
(ii) CMLCEN: Sequential optimization with a central (not-shared) RL policy deciding the precedence indices for each involved robot as a function of features of all the involved robots and pseudo-robots.
(iii) FCFS [9], [16]: First come first serve. Complete trajectory for a robot, say $i \in V$, to cross the intersection is obtained as and when it enters the RoI (at $t_i^{\mathcal{A}}$) considering $V_p(.) = \{i\}$ and $V_s(.) = \{j : j \in V \wedge t_j^{\mathcal{A}} < t_i^{\mathcal{A}}\}$.
(iv) TTR (Time to react) [24]: Sequential optimization with negative of the ratio of distance to intersection and current velocity of a robot as its precedence index.
(v) PDT [33]: Sequential optimization with negative of product of distance to intersection and time-to-react of a robot as its precedence index.
(vi) CDT [25]: Sequential optimization with a convex combination of distance to intersection and time-to-react of a robot as its precedence index, with the convex combination parameter 0.5.
(vii) OCP [28]: Sequential optimization with order in 'virtual' intersection entry and exit times deciding the crossing order. As suggested in [28], these 'virtual' intersection entry and exit times are computed by solving the combined optimization problem (11) at that coordinated phase neglecting the intersection safety constraints (4). Then the steps 1 to 5 proposed in Section III(a) of [28] are followed to decide the crossing order.

We compare the performance of the proposed CML (shared policy) approach against the above mentioned policies for various traffic and parameter settings, for e.g., homogeneous traffic (Sim-1 through Sim-4) and heterogeneous traffic (Sim-5). For Sim-1 through Sim-5, we train different policies for each simulation. We also compare the performance of policies trained on some set of average arrival rates against heuristics on a test set of different average arrival rates (Sim-6 and Sim-7) and time varying average arrival rates (Sim-8 and Sim-9) unseen during training. For testing in Sim-6 to Sim-9, we use the policies trained in Sim-2. These serve as a test for learnt policy generalization. Table 3 presents a comprehensive list of training and testing traffic and other parameters used in different simulations.

We compare the different policies in the following way. We run each of the heuristic policies on 100 randomly generated streams each of $300s$ long, for each average arrival rate. For the learnt policies from CML approach, we run each of the 10 trained policies on 10 randomly generated streams (hence 100 random streams), for each average arrival rate in the simulation (refer to Table 3). We remove the data of first 90s in each stream to neglect transient traffic behaviour and compute the average (over 100 streams) of the control objective values (value of the objective function of Problem (8) with the first $T_h$ seconds of a robot's trajectory data) from generated streams. We call this the *average performance* of policy $p$, denoted by $\bar{J}_p$, for a given arrival rate. We then compare the average performance of the learnt CML policy against some policy $p$ using the quantity

$$E(p) := \frac{\bar{J}_{CML} - \bar{J}_p}{\bar{J}_p}\% \qquad (17)$$

for each average arrival rate in the test cases.

We also compare the learnt policies against the heuristics using weighted average (over all robots over 100 streams) of time to cross (TTC) the intersection since their entry into RoI. For this, say for a given random stream, $\hat{J}_p$ represents the weighted average of the TTC (TTC of each robot weighted by its priority value). Then we compare the learnt CML policy against some policy $p$ using the quantity

$$B(p) := \frac{\hat{J}_{CML} - \hat{J}_p}{\hat{J}_p}\% \qquad (18)$$

for each average arrival rate in the test cases. Both $E(p)$ and $B(p)$ averaged over multiple random streams specify how suboptimal the policy $p$ is compared to the learnt CML policy in terms of distance covered and time taken to cross respectively.

Figures 6 through 11 compare the performance of different heuristics against the policies learnt using CML after 100000 learning iterations under different traffic and parameter configurations. The common legend for Figures 6, 7 and 8 is in Figure 9.

Figures 6 and 7 compare the performance of CML policies over other policies, in homogeneous traffic setting. We see that CML policy, in general, outperforms all the heuristics by a good margin both in terms of average performance (in terms of $E(.)$ and $B(.)$) and average TTC over a large range of average arrival rates. Note that average TTC is an inverse measure of average intersection throughput (low TTC implies high intersection throughput). We also note that the central CML (CMLCEN) policy performs poorly compared to the shared CML policy. This may be due to increase in the number of parameters to be learnt in a central policy and possibly more local minima which come along with it. Since BESTSEQ is computationally expensive, we present comparisons of CML and other heuristics against BESTSEQ only for the set-up in Sim-1 (Figures 6a and 7a), that too for arrival rates 0.01 to 0.05 robots/lane/s. Beyond this arrival rate, the computation time for BESTSEQ is prohibitively large. For very low arrival rates (0.01 to 0.04 robots/lane/s), we notice that the heuristics do better than CML trained policies. This may be because a CML policy generalizes for better performance over a large range of average arrival rates.

In Figure 8 (results related to Sim-6 and Sim-7), we see that the policies trained on homogeneous traffic setting on a range of average arrival rates $\{0.11, 0.12, \ldots, 0.2\}$ outperform other heuristics in test traffic generated using other average arrival rates too. Figures 10a and 11a present comparisons between different policies in heterogeneous traffic setting (Sim-5). In Sim-5, the CML policy outperforms other heuristics by a large margin. This may be due to the bad performance of heuristics in such settings. We see similar results when the same trained policies are tested on time varying traffics – homogeneous burst-mode traffic (Sim-8) presented in Figures 10b and 11b and heterogeneous random traffic (Sim-9) presented in Figures 10c and 11c. Figures 8, 10b, 10c, 11b and 11c also demonstrate that the learnt CML policies generalize well to traffic situations unseen during training.

In Figure 12, we compare the computation times required for getting the crossing order (precedence indices) for various policies/heuristics collected over a random stream of 500s, with average arrival rate for all lanes set to 0.2 robots/lane/s with heterogeneous parameters. Note that computing precedence indices is the only part of the framework where the computation efforts differ for various policies/heuristics, since sequential optimization follows precedence index computations for all heuristics. In Figure 12, we see that heuristics TTR, PDT and CDT take the least time (less than $20\mu s$), the CML policy takes close to 3ms on average and the OCP heuristic takes up to 250ms on average. This can be attributed to the fact that for TTR, PDT and CDT, computations are just algebraic operations. For CML, a neural network needs to be evaluated. However, for OCP, a non-linear optimization problem needs to be solved to get the precedence indices.

Similarly, in Figure 13 we compare the average time taken per robot to solve the provisional and coordinated phase optimization problems for various policies. We run this comparison on 10 (same) random streams on each of the heuristics for Sim-1 and Sim-2 settings. Notice that for CML and other heuristic policies except FCFS, a robot may go through provisional phase multiple times. Due to this, FCFS incurrs far less computation time compared to the heuristics for all arrival rates. For lower arrival rates, CML seems to take the same amount of time as other heuristics, which changes as the arrival rates increase. This may be due to the CML policy allowing for better platooning by making some low priority robots go through more provisional phases compared to other heuristics.

## VI. ADAPTATIONS FOR IMPLEMENTATION

For practical implementation of the algorithm, we need to address the issues arising from the simplifying assumptions on tracking errors, communication delays and computation times made during the formulation in Section II. In the following, we describe how we have relaxed these assumptions to make the framework fit for practical implementation.

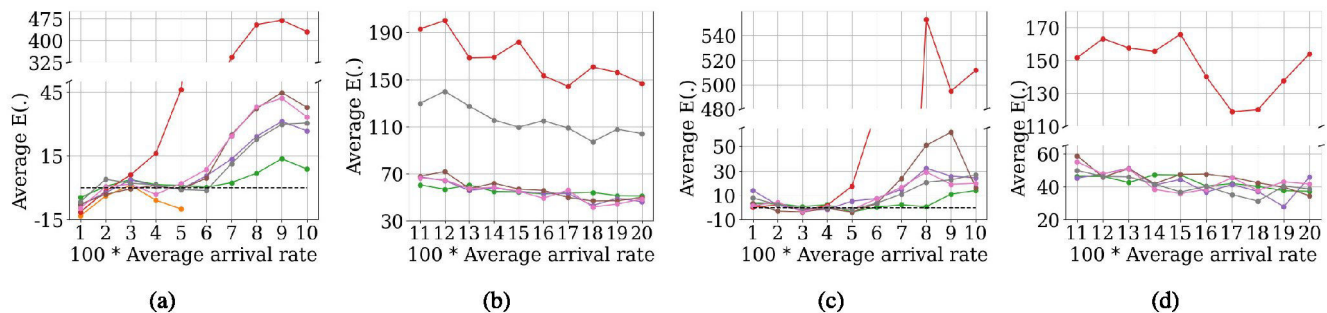

**FIGURE 6.** Percentage improvement in average performance of CML trained policies over that of different heuristics ($E(.)$, see (17)) averaged over 100 random streams for various heuristics and average arrival rates (Figures (a), (b), (c) and (d) for Sim-1, Sim-2, Sim-3 and Sim-4 respectively). Dashed black line is 0%. See Figure 9 for legend.

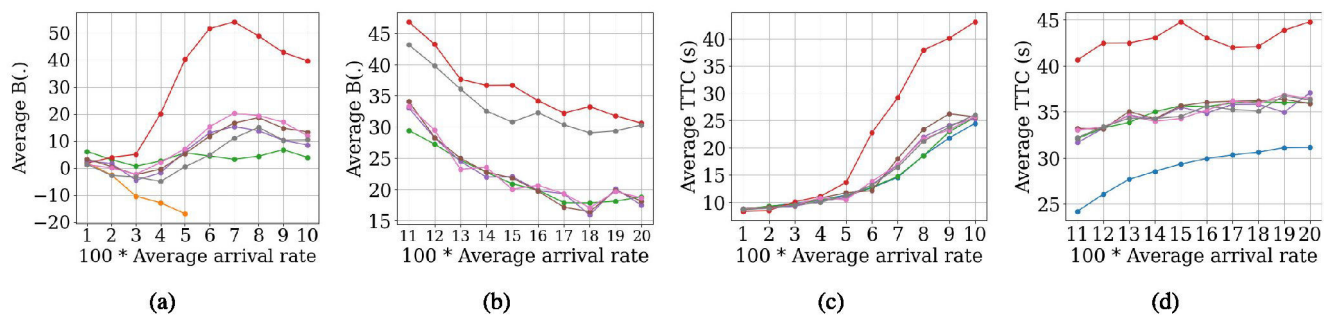

**FIGURE 7.** Percentage reduction in weighted average of TTC for CML when compared with various heuristics ($B(.)$, see (18) and Average time to cross (TTC) for different policies/heuristics averaged over 100 random streams (Figures (a), (b), (c) and (d) for Sim-1, Sim-2, Sim-3 and Sim-4 respectively). See Figure 9 for legend.

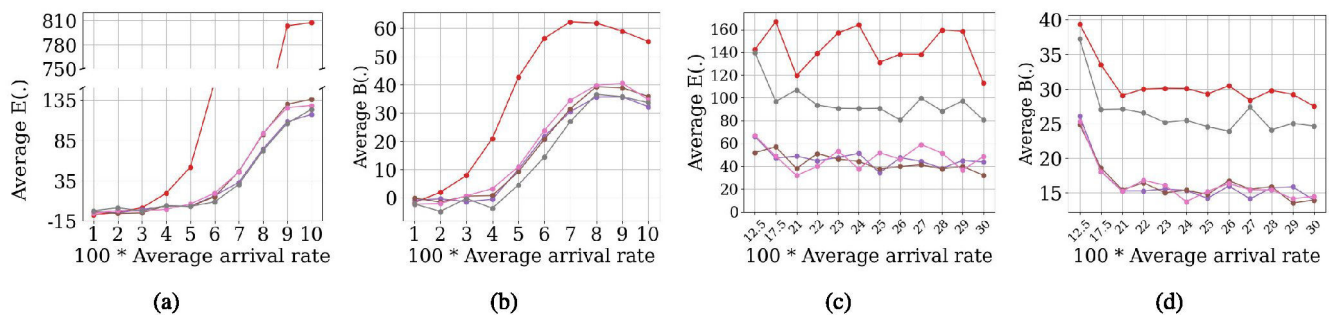

**FIGURE 8.** Percentage improvement in average performance of CML trained policies (average $E(.)$, see (17)) over that of various heuristics averaged over 100 random streams ((a) for Sim-6 and (c) for Sim-7). Comparison of weighted average time to cross (TTC) against CML for various heuristics (average $B(.)$, see (18)) ((b) for Sim-6 and (d) for Sim-7). See Figure 9 for legend.

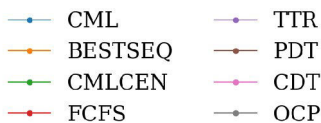

**FIGURE 9.** Legend for Figures 6, 7, 8, 17a and 17b.

We limit the scope of this paper by assuming that a low level trajectory tracking controller is available on each robot which tracks the given reference trajectory faithfully with bounded errors. We also assume that the per-robot communication delay and computation times are bounded.

*Handling tracking errors:*

Suppose that the bound on position tracking errors is $b$ units. The tracking errors are handled by considering $\hat{L}_j = L_j + 2b$ as the length of robot $j \in V$ for all computations in the algorithm. Notice that this affects both rear-end and intersection safety constraints.

*Handling communication and computation delays:*

Suppose that the bound on communication delay is $\delta$ units, bounds on computation time per robot are $\Delta_p$ units and $\Delta_c$ units for provisional phase and coordinated phase, respectively. Computation time and communication delays are handled by pre-computing the trajectories. That is if a robot $i$ is about to enter the RoI at time $t_i^A$, we estimate its velocity of entry and initiate its provisional phase trajectory computations at time $t_i^A - \Delta_p - \delta$. Similarly, we initiate the computations for $k^{th}$ coordinated phase at $kT_c - N_p\Delta_c - \delta$ for the robots in $V_p$ at that time, where $N_p$ is an appropriate upper bound on the number of robots in $V_p$. The computations are made assuming perfect prediction of positions and velocities of robots at $kT_c$.

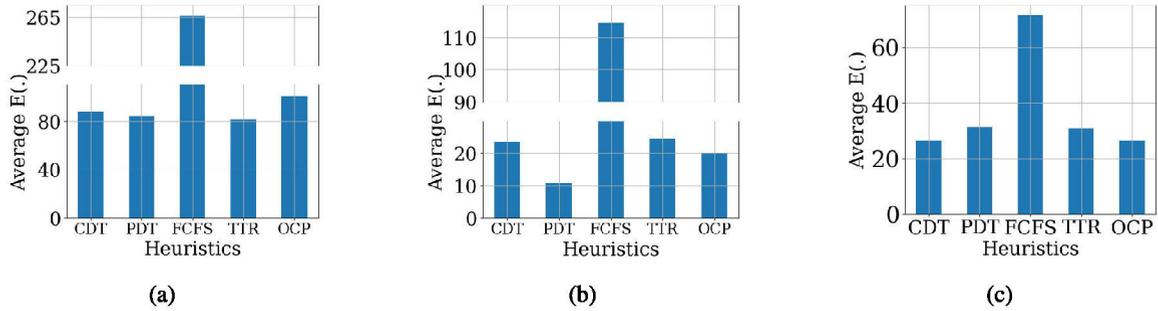

**FIGURE 10.** Percentage improvement in average performance of CML trained policies over that of different heuristics ($E(.)$, see (17)) averaged over 100 random streams for various heuristics and average arrival rates (Figures (a), (b) and (c) for Sim-5, Sim-8 and Sim-9 respectively).

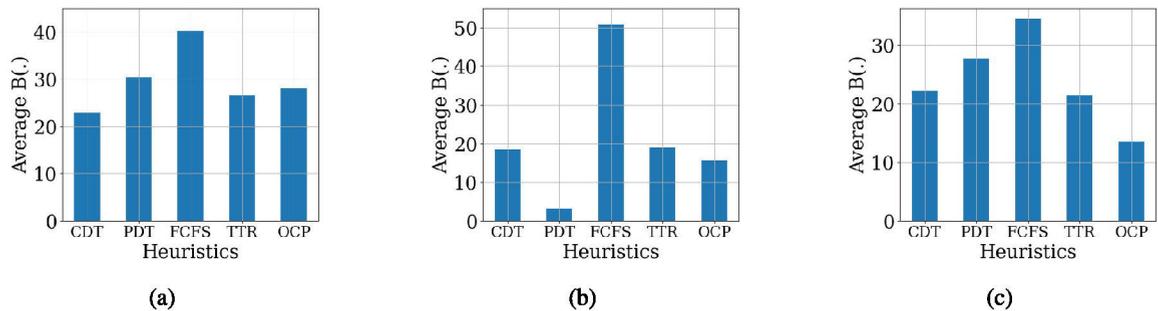

**FIGURE 11.** Comparison of weighted average time to cross (TTC) against CML for various heuristics (average $B(.)$, see (18)) (Figures (a), (b) and (c) for Sim-5, Sim-8 and Sim-9 respectively).

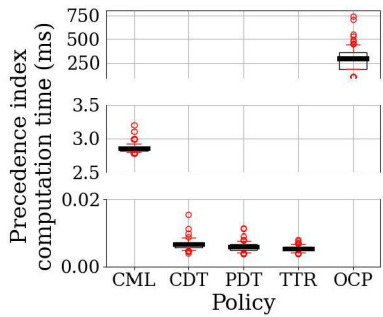

**FIGURE 12.** Comparison of computation times required to get precedence indices (crossing order) for different policies/heuristics.

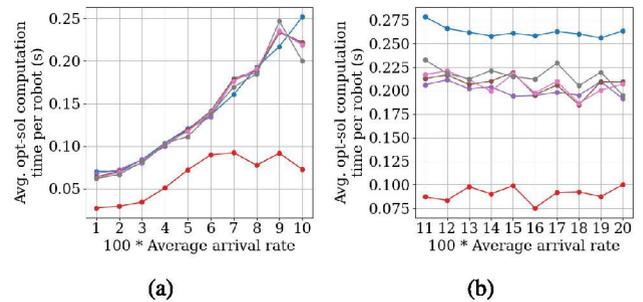

**FIGURE 13.** Average computation time per robot taken to solve provisional and coordinated phase optimization problems for various average arrival rates. Figures (a) and (b) are with simulation parameters corresponding to Sim-1 and Sim-2 respectively. See Figure 9 for legend.

If the communication delays are small enough, the errors due to such delays can be merged with tracking errors by adding the maximum possible distance a robot $i \in V$ can cover during the communication delay period which is $\bar{v}_i\delta$. In this case, trajectory pre-computation times may not involve the communication delay term $\delta$. Since the communication delays are small in our experiments, we follow this method.

Note that the policy is trained offline in an ideal scenario without these adaptations. These adaptations to the algorithm are made only during implementation.

### A. SIMULATION STUDY ON EFFECT OF ADAPTATIONS

Since we train the policies in an ideal scenario and the adaptations are incorporated in the implementation, it is natural to expect some degradation in the performance of the policy in the adapted framework setting. In this regard we compare the performance with and without adaptations for setting in Sim-1 and Sim-2.

Recall that delays due to communication and computation times are handled by pre-computations. Notice that, even with pre-computations, the coordinated phase occurs periodically in the adapted framework with the same period as the ideal framework. Hence, we choose to ignore the computation and communication delays in the following numerical study. Instead, we model any error in prediction of robot state for pre-computations as tracking error.

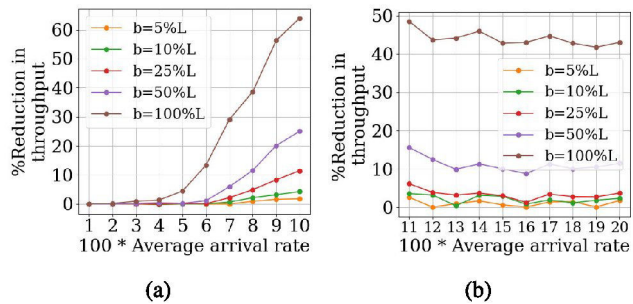

**FIGURE 14.** Percentage reduction in throughput (number of robots crossed per unit time) for various average arrival rates and various values of buffer $b$ when compared against $b = 0$. Figures (a) and (b) are with simulation parameters corresponding to Sim-1 and Sim-2 respectively. $L$ denotes the length of a robot.

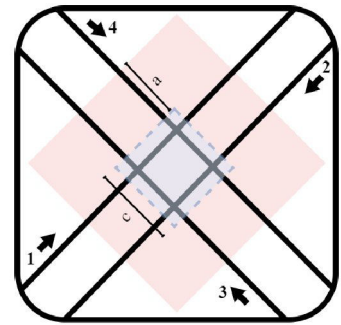

**FIGURE 15.** A schematic of the layout used for implementation on robots. The black lines represent the paths for the robots to follow. The union of red and blue shaded regions is the RoI and the blue shaded region is the intersection. The robots flow in the direction of the arrows depicted along with lane numbers. $a = 0.8$m and $c = 0.53$m.

As noted earlier, tracking errors are addressed by adding buffer $2b$ units to the length of the robot in the constraints. In Figure 14 we study the effect of tracking errors on the average throughput (number of robots crossed per unit time averaged over 10 policies and 10 simulations per policy) by comparing the ideal scenario ($b = 0$) against various values of $b$ for Sim-1 and Sim-2. In all these simulations, we use the CML policies trained on the ideal scenario with $b = 0$. In order to reduce the randomness in the comparisons, we fix the number of robots and the lower bound on their arrival time for all simulations. The actual stream may be different for different values of $b$ due to the effect of the buffer value. In Figure 14, we observe that the average throughput consistently reduces with increase in the buffer value. Also, it is interesting to note that for small buffer values, the reduction in throughput is reasonably small.

## VII. IMPLEMENTATION IN A LAB SETTING

We implemented the proposed algorithm on a collection of line following robots, specifically 3pi+ 32U4 Turtle edition robots manufactured by Pololu Robotics and Electronics [65]. Due to the lack of significant computational and communication capabilities on these robots, we run the algorithm on a computer, store the computed trajectories and use XBee S6B WiFi modules to communicate the trajectories to the robots at each time-step. We use OptiTrack motion capture system [66] to track these robots. We use a 4-way intersection layout as in Figure 15 printed on a flex-sheet. The black lines represent the lanes for the robots. The robots follow the lines to cross the intersection and loop around on the perimeter square and re-enter the RoI on a randomly chosen lane. This process continues repeatedly so that we have a continual robot streams.

### A. PARAMETERS FOR EXPERIMENTS

The approach length for each lane in the RoI, $d(l) = 0.8$m (depicted as $a$ in Figure 15) $\forall l \in \{1, 2, 3, 4\}$ and the width of the intersection in 0.53m (depicted as $c$ in Figure 15). The length of each robot is 0.1m. The bound on inherent tracking errors and tracking errors due to communication delay was measured to be, $b = 0.075$m. To allow for better trajectory tracking performance, we set $\bar{v}_i = 0.25$m/s $\forall i \in V$, even though the robots are capable of speeds up to 0.4m/s. Given these bounds, we chose $\hat{L}_j = 0.15$m for buffer to address communication delays and tracking errors. We set $T_c = 3$s and $T_h = 20$s. We measured $\Delta_p = 0.1$s and $\Delta_c = 0.2$s after repeated experiments. We use a PID position tracking controller for low-level trajectory tracking, which faithfully tracks the trajectory with bounded error. Figure 16 shows a picture of the actual lab setup where we ran these experiments.

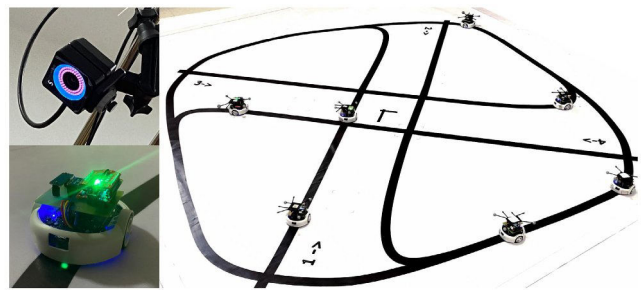

**FIGURE 16.** Setup for hardware implementation in our lab. The bottom-left picture is a close-up of a robot used, the top-left picture is one of the cameras of OptiTrack motion capture system. The picture on the right depicts an ongoing experiment.

### B. INDICATIVE RESULTS FROM EXPERIMENTS

We deployed a policy that was learnt offline in ideal simulations (without tracking errors, communication and computation delays and restricting parameters only to the RoI) using CML approach. We trained a set of 10 policies each with different network parameter initialization on the set of average arrival rates $\{0.11, 0.12, \ldots, 0.2\}$ robots/lane/s. We tested these 10 learnt policies in ideal simulation environment and chose the policy with the highest sum of average objective function value over the average arrival rates in the set $\{0.01, 0.02, \ldots, 0.2\}$ robots/lane/s, over 10 test simulations for each average arrival rate and compare its performance against the heuristics. The plots indicating the results of these simulations is presented in Figures 17a and 17b. We observe that, for the parameters considered in the experiments, the CML policies start to outperform the other heuristics only after an average arrival rate of 0.1 robots/lane/s.

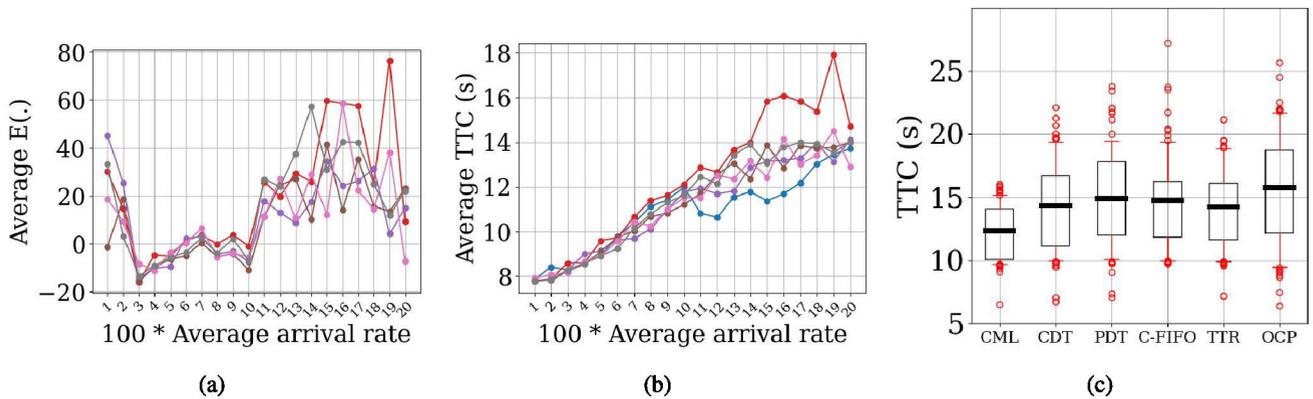

**FIGURE 17.** Figure (a): Percentage improvement in average performance of CML trained policy over that of different policies and Figure (b): Average time to cross (TTC) for different policies. Both the plots are under ideal simulation environment for parameters used in experiments. See Figure 9 for legend. Figure (c): Box plot of TTC values for 3 runs of 3 different initial positions each for different policies. The box represent the range of values between 1st and 3rd quartile, whiskers represent range of values between 10th and 90th percentile and bold black line represents the mean of the TTC values for each policy.

However, due to the limitations imposed in our lab setting (e.g., number of robots, approach length for each lane, velocity bounds on the robots etc.), it is hard to achieve such high average arrival rates consistently. For this reason, we choose to present some indicative experimental results for snapshots (sets $V_p$ and $V_s$).

In this regard, we consider the following 3 different robot initializations on the outer ring (part of the path outside the RoI in Figure 15):

(i) 2 robots to enter each lane.
(ii) 3 robots to enter two conflicting lanes and 1 robot each to enter the other two lanes.
(iii) 2 robots to enter two conflicting lanes, 3 robots to enter one of the other and 1 robot to enter the other.

The performance of the learnt policy is compared in the real-world set-up with the algorithmic adaptations proposed in the previous subsections against other heuristics by comparing the time to cross values such that each of the involved robots got through coordinated phase exactly once. The value of $T_c = 5s$ is chosen so that more robots participate in a coordinated phase. Note that the FCFS heuristic has an undue advantage in such a comparison since the trajectories are computed for a robot as and when it enters the RoI and does not have to wait till the next coordinated phase. Hence, to be fair for the heuristics, here we propose to compare with an alternate heuristic which we call *C-FIFO*, where the robots go through provisional phase and the crossing order (and hence the order in coordinated phase) follows the first-in first-out rule.

We conduct 3 runs for each initialization for each policy (each run with same initial position of the robots) and present the average TTC obtained for different policies in Figure 17c.

In Figure 17c we observe that the CML policy produces low TTC values compared to other heuristics. We also observe that most of the TTC values for CML policy lie in a small region (e.g. range of values inside the box i.e., which fall within 1st and 3rd quartile) compared to other heuristics, thus promoting fairness in TTC among the involved robots in various scenarios. This is indicative of the performance of the CML policy deployed under real-time constraints.

As mentioned earlier, due to restrictions imposed in the lab-setting, we do not present results from our experiments with longer continual streams of robots. A short video of our lab implementation with continual streams negotiating the intersection safely can be seen using the link https://youtu.be/Io4DxmpJPaI. This stands as a proof of concept for real-time implementability of the methods proposed in this work. We leave the larger scale implementation for future work.

## VIII. CONCLUSION AND FUTURE WORK

In this paper, we have combined learning with optimization methods to obtain a near-optimal solution for multi-robot unsignalized intersection management, which can be implemented in real-time. The proposed solution gives a policy that is shared by all the robots and can be deployed in a distributed manner. With extensive simulations, we have established that such a policy outperforms the major heuristics proposed in the literature, over a range of average traffic arrival rates. We also illustrate the vast improvement in computation time of the proposed solution compared to that of naive optimization methods for the intersection management problem. We have also proposed some adaptations to the solution framework to address real-world challenges like tracking errors, communication and computation delays and have implemented the learnt policies on robots in a lab setting with the proposed adaptations. The proposed method is flexible so that, with fresh training, policies for different kinds of intersections can be learnt. As already seen, such a training will need some basic information about the new intersection setting and no expert knowledge is necessary (as in the case of some heuristics and analytical methods). One future research direction is to use transfer learning methods to reduce training time when a policy learnt for one intersection setting is to be adapted to another setting. It is also interesting to see if a single policy can be learnt to work for a set of intersections and robots with different parameters like lane and RoI dimensions, robot velocity and acceleration limits etc. Other future directions include implementation fine-tuning, extensions to allow lane changes, to handle disturbances, dynamic obstacles and for a network of intersections.

## ACKNOWLEDGMENT

The authors would like to thank Ayush Das, Rithvik Mahajan, and Soumyodipta Nath for their help in setting up and implementing the algorithm on robots in the lab setting.

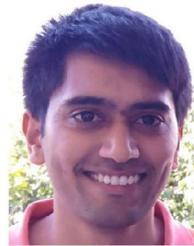

**NISHCHAL HOYSAL G** (Student Member, IEEE) received the B.E. degree in mechanical engineering from the B.M.S. College of Engineering, Bengaluru, India, in 2016. He is currently pursuing the Ph.D. degree with the Robert Bosch Centre for Cyber Physical Systems, Indian Institute of Science at Bengaluru.

His research interests include learning methods for multi-agent planning and coordination.

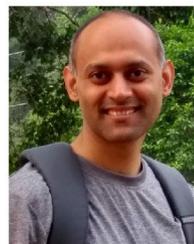

**PAVANKUMAR TALLAPRAGADA** (Member, IEEE) received the B.E. degree in instrumentation engineering from the SGGS Institute of Engineering and Technology, Nanded, India, in 2005, the M.Sc. (Engg.) degree in instrumentation from Indian Institute of Science at Bengaluru, in 2007, and the Ph.D. degree in mechanical engineering from The University of Maryland, College Park, in 2013. He was a Postdoctoral Scholar with the Department of Mechanical and Aerospace Engineering, University of California at San Diego, San Diego, from 2014 to 2017. He is currently an Associate Professor with the Department of Electrical Engineering and the Robert Bosch Centre for Cyber Physical Systems, Indian Institute of Science at Bengaluru. His research interests include networked control systems, distributed control, multi-agent systems, and dynamics in social networks.

● ● ●